\documentclass[10pt,twocolumn,letterpaper]{article}

\usepackage{iccv}
\usepackage{times}
\usepackage{epsfig}
\usepackage{graphicx}
\usepackage{amsmath}
\usepackage{amssymb}

\usepackage{graphicx}
\usepackage{amsmath}
\usepackage{amssymb}
\usepackage{booktabs}

\usepackage{algpseudocode}
\usepackage{algorithm}
\usepackage{array}
\usepackage{multirow}
\usepackage{dsfont}
\usepackage[font={small}]{caption}
\usepackage{wrapfig,lipsum,booktabs}

\usepackage{color}
\usepackage{xcolor}
\definecolor{citecolor}{HTML}{0071bc}
\usepackage{xspace}

\definecolor{citecolor}{HTML}{0071bc}
\usepackage[pagebackref=true,breaklinks=true,colorlinks,citecolor=citecolor,bookmarks=false]{hyperref}

\usepackage[capitalize]{cleveref}
\crefname{section}{Sec.}{Secs.}
\Crefname{section}{Section}{Sections}
\Crefname{table}{Table}{Tables}
\crefname{table}{Tab.}{Tabs.}

\ificcvfinal\fi

\newlength\savewidth\newcommand\shline{\noalign{\global\savewidth\arrayrulewidth\global\arrayrulewidth 1pt}\hline\noalign{\global\arrayrulewidth\savewidth}}

\newcommand{\IGNORE}[1]{}

\usepackage[breaklinks=true,bookmarks=false]{hyperref}

\iccvfinalcopy 

\begin{document}

\title{ActorsNeRF: Animatable Few-shot Human Rendering with \\ Generalizable NeRFs}

\author{Jiteng Mu\textsuperscript{1}, \quad
Shen Sang\textsuperscript{2}, \quad
Nuno Vasconcelos\textsuperscript{1}, \quad
Xiaolong Wang\textsuperscript{1} \quad \\
\textsuperscript{1}UC San Diego, \textsuperscript{2}ByteDance
}


\twocolumn[{
\vspace{-1em}
\maketitle
\vspace{-1em}

\begin{center}
    \centering
    \includegraphics[width=\linewidth]{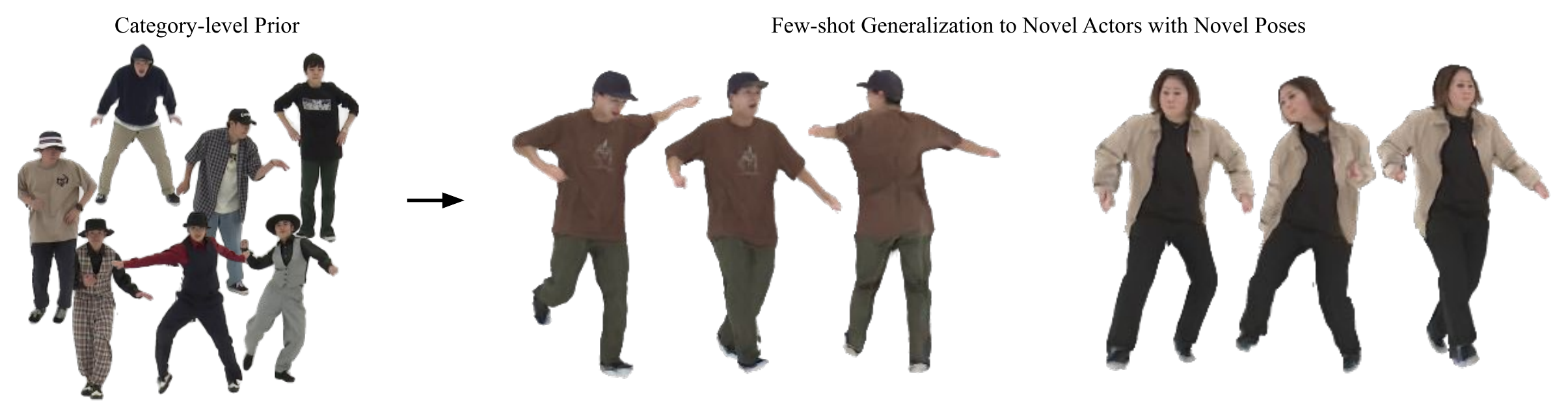}
    \captionof{figure}{
    Animatable NeRF from a few images. We present ActorsNeRF, a category-level human actor NeRF model that generalizes to unseen actors in a few-shot setting. With only a few images, e.g., 30 frames, sampled from a monocular video, ActorsNeRF synthesizes high-quality novel views of novel actors in the AIST++ dataset with unseen poses (shown on the right).
    }
    \label{fig:teasing}
\end{center}
}]

\begin{abstract}
    While NeRF-based human representations have shown impressive novel view synthesis results, most methods still rely on a large number of images / views for training. In this work, we propose a novel animatable NeRF called ActorsNeRF. It is first pre-trained on diverse human subjects, and then adapted with few-shot monocular video frames for a new actor with unseen poses. Building on previous generalizable NeRFs with parameter sharing using a ConvNet encoder, ActorsNeRF further adopts two human priors to capture the large human appearance, shape, and pose variations. Specifically, in the encoded feature space, we will first align different human subjects in a category-level canonical space, and then align the same human from different frames in an instance-level canonical space for rendering. We quantitatively and qualitatively demonstrate that ActorsNeRF significantly outperforms the existing state-of-the-art on few-shot generalization to new people and poses on multiple datasets. Project page: \href{https://jitengmu.github.io/ActorsNeRF/}{https://jitengmu.github.io/ActorsNeRF/}. 
\end{abstract}


\section{Introduction}

Recent advances in Neural Radiance Fields (NeRF)~\cite{Su21anerf} have enabled significant progress in free-viewpoint rendering of humans performing complex movements. The possibility of achieving photo-realistic rendering is of major interest for various real-world applications in AR or VR. However, to achieve high-quality rendering, existing approaches~\cite{Peng2021neuralbody,Liu21neuralactor,Peng21animatablenerf,Noguchi21narf,Su21anerf} require a combination of synchronized multi-view videos and an instance-level NeRF network, trained on a specific human video sequence. While results are encouraging, the multi-view requirement is a significant challenge to applications involving videos in the wild. Recently, progress has been made to eliminate this constraint, by enabling human  rendering from a monocular video~\cite{Weng2022humannerf,Jiang22neuman}. However, these approaches still require a large number of frames, which covers a person densely from all viewpoints.

In this work, we consider the more practical setting and ask the question: Can an animatable human model be learned from just a few images? We hypothesize that this is possible by introducing  a class-level encoder, trained over multiple people, as shown in Figure~\ref{fig:teasing}. 
This hypothesis has been demonstrated by recent works on generalizable NeRFs~\cite{Yu2021pixelnerf,Chen21mvsnerf}, where an encoder network is trained across multiple scenes or objects within the same category to construct NeRF. By parameter sharing through the encoder, the  prior learned across different scenes can be re-used to perform synthesis even with a few views. However, most approaches can only model static scenes.
We investigate how generalizable NeRF can be extended to the learning of a good prior for the much more complex setting of videos of humans performing activities involving many degrees of freedom, large motions, and complex texture patterns.

For this, we introduce \textbf{ActorsNeRF}, a category-level human actor NeRF model that is transferable, to unseen humans in novel action poses, in a few-shot setup. This setup requires the animation of a previously unseen human actor into unseen views and poses, from a few frames of monocular video. Such generalization requires more than simple parameter sharing via an encoder network, and can benefit from the incorporation of explicit human priors. Our insight is that, while human actions and appearances are complex, all humans can be {\it coarsely\/} aligned in a \emph{category-level canonical space} using a parametric model such as SMPL~\cite{Loper2015SMPL}. {\it Fine-grained\/} alignment can then benefit from an \emph{instance-level canonical space} derived from both this prior and the few-shot data available for the target actor.

To implement this insight, we endow ActorsNeRF with a 2-level canonical space. Given a body pose and a rendering viewpoint, a sampled point in 3D space is first transformed into a canonical (T-pose) space by linear blend skinning (LBS)~\cite{Loper2015SMPL}, where the skinning weights are generated by a skinning weight network that is shared across various subjects. Since LBS only models a coarse shape (similar to an SMPL mesh), we refer to this T-pose space as the {\it category-level canonical space.\/} Direct rendering from the latter fails to capture the shape and texture details that distinguish different people. To overcome this limitation, points in the category-level canonical space are further mapped into an {\it instance-level canonical space\/} by a deformation network. A rendering network finally maps the combination of pixel-aligned encoder features and points into corresponding colors and densities.

ActorsNeRF is designed such that the combination of feature encoder and skinning weight network forms a category-level shape and appearance prior, and the deformation network learns the mapping to the instance-level canonical space. To adapt to a novel human actor at test time, only the deformation network (instance-level) and rendering network are fine-tuned with the few-shot monocular images. The image encoder and skinning weight network are frozen.

We quantitatively and qualitatively demonstrate that ActorsNeRF outperforms the existing approaches by a large margin on various few-shot settings for both ZJU-MoCap Dataset~\cite{Peng2021neuralbody} and AIST++ Dataset~\cite{Li21AIST}.
To the best of our knowledge, we are the first to explore few-shot generalization from few-shot monocular videos in the context of NeRF-based human representations. 





\section{Related Work}

\begin{figure*}
    \centering
    \includegraphics[width=0.95\linewidth]{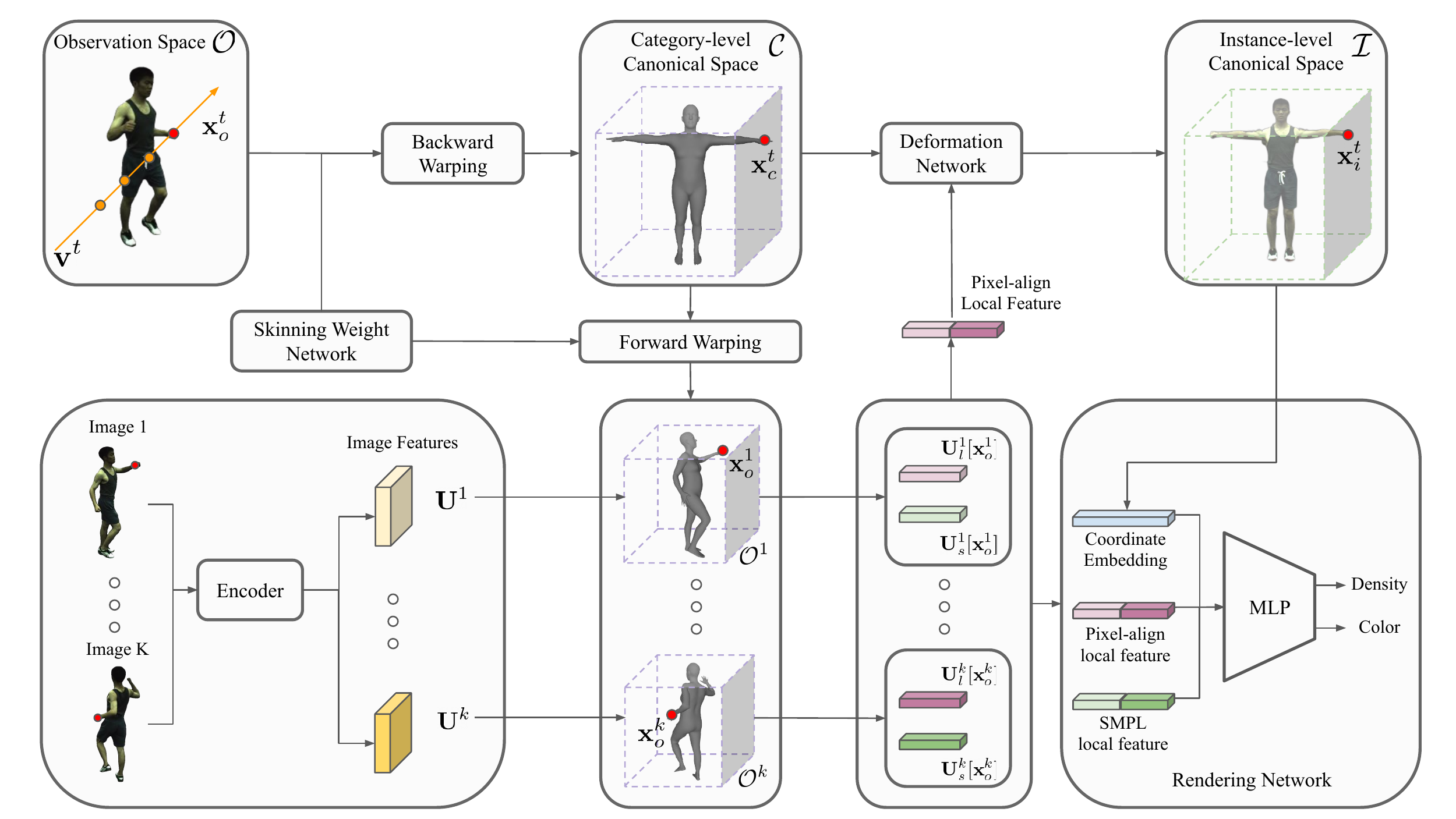}
    \vspace{-0.10in}
    \caption{Overview of ActorsNeRF. First, $K$ images are passed through an encoder to extract feature tensors $\mathbf{U}_{l}^{k}$. Given a target pixel location along with a view direction $\mathbf{v}^t$, a point $\mathbf{x}_o^t$ is sampled along the ray. The point is then mapped to the category-level canonical space $\mathbf{x}_c^t$ through a backward warping. Then, $\mathbf{x}_c^t$ is transformed to $K$ other observation spaces through corresponding forward warpings and projected to images to query corresponding local features (pixel-aligned features $\mathbf{U}_{l}^{k}$ and SMPL local features $\mathbf{U}_{s}^{k}$). Next, a deformation network takes both $\mathbf{x}_c^t$ and its pixel-aligned features to produce a location $\mathbf{x}_i^t$ in the instance-level canonical space. $\mathbf{x}_i^t$ along with its pixel-aligned features are then mapped to color and density for volume rendering.}
    \label{fig: overview}
    \vspace{-0.10in}
\end{figure*}


\textbf{Dynamic NeRF.} While NeRF~\cite{mildenhall2020nerf} was originally proposed for modeling static scenes, recent efforts have successfully extended it to dynamic scenes~\cite{Xian21spacetime,Li21neuralsceneflow,Gao21dynamicviewsynthesis,Park21nerfie,Park21hypernerf} and deformable objects~\cite{Wei22selfsupervised,Pumarola21dnerf,Kania22conerf,Raj21pva, Yang22banmo,Corona22lisa,Fang22TiNeuVox,guo22neuraldeformablevoxel}. One strategy to model dynamic objects and scenes is to align observations from various time steps in a canonical space, decoupling 4D as 3D and time reduces the  complexity. For example, Pumarola et al.~\cite{Pumarola21dnerf} propose a warping field to map sampled points to template space and then directly render from the canonical space. Our work shares the principle of aligning different observations for efficient modeling. Going beyond an instance canonical space, ActorsNeRF also incorporates the category-level human prior in the modeling of unseen actors with few images.


\textbf{NeRF-based Human Rendering.} Human-specific rendering is a longstanding challenge~\cite{Kanade97virtualizedreality,Matusik00imagebased,Ye13freeviewpoint,Collet15highquality,Dou16fusion4d,Sarkar21neuralrerender,Grigorev19coordinate,BruallaP18lookinggood,Casas14video,Sarkar21style} due to the large modeling space of shape, pose, and appearance. Recently, NeRF-based human representations have shown promise for high-quality view synthesis~\cite{Peng2021neuralbody,Peng21animatablenerf,Xu21HNeRF,Noguchi21narf,Kwon21nhp,Liu21neuralactor,Su21anerf,Weng2022humannerf,Zhao22humannerf,Jiang22neuman,Gao22mpsnerf,Li22tava,Wang22ARAH,Shao22doublefield,su22danbo}. For example, Peng et al.~\cite{Peng2021neuralbody} propose to attach a set of latent codes to SMPL~\cite{Loper2015SMPL} and render novel views of a performer from sparse multi-view videos. To better animate the human actor, subsequent works~\cite{Liu21neuralactor,Peng21animatablenerf} introduce a canonical space to align different body poses. These methods require multiview videos, which limits their application. To address this challenge, Weng et al.~\cite{Weng2022humannerf} further decompose shape and pose into skeletal motion and non-rigid shape deformation, and synthesize photorealistic details from just a monocular video. Jiang et al.~\cite{Jiang22neuman} jointly learns a human NeRF and a scence NeRF from a monocular video. However, these methods are limited to instance-level and does not generalize to novel human actors. To achieve better generalization, category-level  NeRFs~\cite{Kwon21nhp,Zhao22humannerf,Gao22mpsnerf} are introduced but these models require multi-view images for both training and inference. Different from all previous approaches, ActorsNeRF learns a category-level generalizable NeRF that allows for novel pose animation on unseen humans, only requiring few-shot images from a monocular video during inference.

\textbf{Few-shot NeRF.} NeRF generally suffers from sub-optimal solutions when trained from a few views. To address this problem, various regularizations~\cite{Jain21dietnerf,Roessle22densedepthprior,Deng22depthsupervised,Niemeyer22regnerf,Chen22geoaug} have been proposed for the few-shot setting. For example, Jain et al.~\cite{Jain21dietnerf} leveraged the pre-trained CLIP~\cite{Radford21clip} model and enforced semantic consistency in the feature space. Other works proposed to avoid the degenerate solutions by introducing a stronger geometry prior, using either supervised~\cite{Roessle22densedepthprior,Deng22depthsupervised} or unsupervised~\cite{Niemeyer22regnerf,Chen22geoaug} depth information. Instead of designing priors, a data-driven way to achieve few-shot transfer, is to train generalizable NeRFs on a large-scale dataset~\cite{Wang21ibrnet,Chen21mvsnerf,Yu2021pixelnerf,Tancik21learninit}. Specifically, an encoder is trained to learn the data prior, and the model will be adapted on an unseen example by fine-tuning. We see ActorsNeRF as an integration of the data-driven approach with large-scale data and the regularization manner using human-specific priors.


\section{Method} \label{method: method}

We introduce ActorsNeRF, a category-level generalizable NeRF model capable of synthesizing unseen humans with novel body poses in a few-shot setup. In order to achieve generalization across different individuals, a category-level NeRF model is first trained on a diverse set of subjects. During the inference phase, we fine-tune the pre-trained category-level NeRF model using only a few images of the target actor, enabling the model to adapt to the specific characteristics of the actor.

Mathematically, the goal is to, given the small set of $M$ frames $I = \{ I^m\}$ capturing a human actor with random poses, learn the parameters $\theta$ of a network $\mathcal{Q}$ that maps a sampled $3D$ point $\mathbf{x} \in \mathbb{R}^3$ into a color vector $\mathbf{c} = (r,g,b)$ and a density $\sigma$, for any randomly sampled rendering viewpoint $\mathbf{v} \in \mathbb{R}^3$ and target pose $\mathbf{S}_{P} \in \mathbb{R}^{(24 \times 3)}$,
\begin{equation}
    (\sigma, \mathbf{c}) = \mathcal{Q}(\mathbf{x}, \mathbf{v}, \mathbf{S}_{P}, I; \theta).
    \label{eq: Q mapping}
\end{equation}

This is quite challenging when only a few images are provided, e.g., $M=5$. To address this problem, we propose two key ideas: 1) leveraging a large-scale dataset of monocular videos to learn a category-level prior for the mapping such that the model can quickly adapt to the new person in a few-shot setting, while 2) incorporating human-specific knowledge to align people of diverse shapes and body poses, using a two-level canonical space.

\subsection{ActorsNeRF} \label{method: ActorsNerf}

To learn the mapping of Eq.(\ref{eq: Q mapping}), at a high-level, we assume having access to a training set of multiple monocular videos capturing different people to learn a category-level prior. Then the model is finetuned on M frames to adapt to the new person.

For both category-level pre-training and finetuning, our idea is to first map a sampled point to a category-level canonical space and then an instance-level canonical space sequentially, where the human body is represented in the canonical pose of Figure~\ref{fig: overview}, and then rendered to color and density conditioned on the encoder features. Mathematically, we define the 3D space associated with a frame of the actor as {\it observation space\/} ${\cal O} \subset \mathbb{R}^3$, the corresponding category-level canonical space $\mathcal{C} \subset \mathbb{R}^3$, the corresponding instance-level canonical space $\mathcal{I} \subset \mathbb{R}^3$. We then define, a forward warping $\mathcal{T}: \mathcal{C} \rightarrow \mathcal{O}$ and a backward warping $\mathcal{T}^{-1} : \mathcal{O} \rightarrow \mathcal{C}$. In addition, $K$ out of $M$ images in the support set are randomly sampled, and an encoder $\mathcal{E}$ is used to obtain the corresponding feature tensor $\mathbf{U}^k$ (for extracting local features).

As shown in Figure~\ref{fig: overview}, to render a target image, a pixel and target rendering viewpoint $\mathbf{v}^t$ define a ray of points $\mathbf{x}^t_o \in \mathcal{O}^t$. Each point is first mapped to a point $\mathbf{x}_c^t$ in the category-level canonical space. The transformation is through a {\it backward warping} $\mathcal{T}^{-1}$, guided by {\it skinning weight network\/} $\mathcal{W}$.

To adapt the canonical, category-level, shape to the shape details of different human actors, a {\it deformation network\/} $\mathcal{D}$ is implemented to transform $\mathbf{x}_c^t$ to a location $\mathbf{x}_i^t$ in the {\it instance-level canonical space\/} $\mathcal{I}$, where the warping is guided by image local features extracted from a set of $\mathbf{U}^{k}$. To extract the pixel-aligned features, the category-level canonical space point $\mathbf{x}^t_c$ is first mapped by corresponding {\it forward warpings} $\mathcal{T}$, into observation spaces $\mathcal{O}^k$ ($k \neq t$) corresponding to support set image $I^k$, and then projected to corresponding feature maps. 

Finally, using $\mathbf{x}_i^t$ and the image local features, a NeRF-based rendering network $\mathcal{R}$ outputs its color and density. We next detail each component of ActorsNeRF, consisting of image encoder $\mathcal{E}$, skinning weight network $\mathcal{W}$, deformation network $\mathcal{D}$, and rendering network $\mathcal{R}$.


\textbf{Feature Encoder $\mathcal{E}$.}  Prior works~\cite{Yu2021pixelnerf,Chen21mvsnerf,Kwon21nhp,Zhao22humannerf} have shown that encoder features learned at a category-level  greatly improve NeRF generalization. We use an encoder $\mathcal{E}$, e.g., ResNet-18~\cite{He16resnet}, to extract a feature tensor $\mathbf{U}^k = \mathcal{E}(I^k)$ for each support set image $I^k$. From $\mathbf{U}^k$, a set of local features $[\mathbf{U}^k_l, \mathbf{U}^k_s]$ is then produced as follows. Pixel-aligned local features $\mathbf{U}^k_l$ are obtained by extracting local features from the feature tensor $\mathbf{U}^{k}$ aligned with each pixel. Specifically, a 3D point $\mathbf{x} \in \mathcal{O}^k$, subject to the camera mapping $\Pi^k$, has pixel-aligned local features $\mathbf{U}^k_l[\mathbf{x}] = interp(\mathbf{U}^k(\Pi^k(\mathbf{x})))$, where $interp$ denotes bi-linear interpolation. On the other hand, SMPL local features $\mathbf{U}^k_s$ are pixel-aligned local features localized by a SMPL model~\cite{Loper2015SMPL}. Let $\mathbf{s}^k_j \in \mathbb{R}^3, j= \{ 1,\cdots, 6890\},$ be the vertices of a SMPL model fitted to  $I^k$. Each vertex is assigned a local feature $\mathbf{U}^k_s[\mathbf{s}^k_j] = interp(\mathbf{U}^k(\Pi^k(\mathbf{s}^k_j)))$.

\textbf{Skinning Weight Network $\mathcal{W}$.} The skinning weight network generates the linear blend of skinning weights for different individuals in the category-level canonical space, which enables the transformations between the category-level canonical space and observation spaces. The design of the network follows~\cite{Weng2022humannerf}, given $B$ ($B=24$ in this paper) joints defined on the human body, the linear blend skinning weights in the category-level canonical space $\mathcal{C}$ are represented by a 3D volume. The difference is that, in ActorsNeRF, network parameters are shared across all actors in the training set to capture the category-level shape prior. An operator $\mathbf{W}[\cdot]$ is defined for tri-linear interpolation of the feature volume $\mathbf{W}$.

\textbf{Forward and Backward Transformation.}
Aggregating image features from the same actor under various body poses requires identifying correspondences between the matching body points $\mathbf{x}_o^k$ of different observation spaces $\mathcal{O}^k$. The common representation through point $\mathbf{x}_c$ in canonical space  $\mathcal{C}$ enables this, by introduction of forward $\mathcal{T}: \mathcal{C} \rightarrow \mathcal{O}$ 
and a backward  $\mathcal{T}^{-1} : \mathcal{O} \rightarrow \mathcal{C}$ warpings. 

Given body pose $\mathbf{S}_P$, a transformation set $\mathbf{T}(\mathbf{S}_P) = \{ \mathbf{T}_1, \cdots, \mathbf{T}_B \}$ is computed for each of the $B$ joints. Location $\mathbf{x}_c \in \mathcal{C}$ is then mapped into point $\mathbf{x}_o  \in \mathcal{O}$ by linear blend skinning (LBS)~\cite{Loper2015SMPL}, 
\begin{equation}
    \mathbf{x}_o = \mathcal{T}(\mathbf{x}_c) = \biggl( \sum_{b=1}^{B}\mathbf{W}^b[\mathbf{x}_c] \mathbf{T}_b \biggr) \mathbf{x}_c
    \label{eq: forward transformation weights}
\end{equation}
where $\mathbf{W}^b[\mathbf{x}_c]$ denotes the $b$th channel of the sampled blending weights at location $\mathbf{x}_c$.

Similarly, the backward mapping $\mathcal{T}^{-1}$ is defined as,
\begin{equation}
    \mathbf{x}_c = \mathcal{T}^{-1}(\mathbf{x}_o) = \biggl( \sum_{b=1}^{B}\mathbf{W}_o^b[\mathbf{x}_o] \mathbf{T}_b^{-1} \biggr) \mathbf{x}_o
    \label{eq: backward transformation}
\end{equation}
where $\mathbf{W}_o^b[\mathbf{x}_o]$ denotes the $b$th channel of the sampled observation-space blending weights for point $\mathbf{x}_o$, given by\cite{Weng2022humannerf,Weng20vid2actor}
\begin{equation}
    \mathbf{W}_o[\mathbf{x}_o] = \frac{\mathbf{W}[\mathbf{T}^{-1}_b \mathbf{x}_o]}{\sum_{b=1}^{B}\mathbf{W}[\mathbf{T}^{-1}_b \mathbf{x}_o]}
    \label{eq: backward transformation}
\end{equation}

As shown in Figure~\ref{fig: overview}, given a point $\mathbf{x}_o^t$ in observation space $\mathcal{O}^t$, corresponding points $\mathbf{x}_o^k$ in support set observation spaces $\mathcal{O}^k$ can be established by first backward mapping $\mathbf{x}_o^t$ to the category-level canonical space point $\mathbf{x}_c^t$ and then forward mapping this point to the points $\mathbf{x}_o^k$. These mappings are key to allow the query of pixel-aligned features from support set images capturing different body poses, without requiring a multi-view imaging setup.

\textbf{Deformation Network $\mathcal{D}$.}
To compensate for the missing details in the category-level canonical space, a deformation network $\mathcal{D}$, parameterized by $\theta_{\mathcal{D}}$, is implemented to transform a point $\mathbf{x}_c$ to a point $\mathbf{x}_i$ in a fine-grained instance-level canonical space $\mathcal{I} \subset \mathbb{R}^3$. This deformation is conditioned on the target body pose $\mathbf{S}_p$ and $K$ support set pixel-aligned local features $\mathbf{U}_l^k$,
\begin{equation}
    \mathbf{x}_i = \mathcal{D}(\mathbf{x}_c, \{ \mathbf{U}_l^k \}_{k=1}^{K}, \mathbf{S}_p ; \theta_{\mathcal{D}}).
    \label{eq: deformation network}
\end{equation}

\textbf{Rendering Network $\mathcal{R}$.}
A rendering network parameterized by $\theta_{\mathcal{R}}$ then predicts the color and density for an instance-level 3D location $\mathbf{x}_i$,
\begin{equation}
    (\sigma, \mathbf{c}) = \mathcal{R}(\mathbf{x}_i, \{ \mathbf{U}_l^k \}_{k=1}^{K}, \{ \mathbf{U}_s^k \}_{k=1}^{K}; \theta_{\mathcal{R}}),
    \label{eq: rendering network}
\end{equation}
conditioned of the sets of pixel-aligned local features $\{ \mathbf{U}_l^k \}_{k=1}^{K}$ and SMPL local features $\{ \mathbf{U}_s^k \}_{k=1}^{K}$. One problem is that, since $\mathbf{U}_s^k$ are only defined on SMPL vertices, features can not be directly queried for sampled points in a continuous space. Therefore, we expand SMPL features to continuous space as follows: K SMPL local features $\mathbf{U}_s^k$ are first concatenated and then passed through a sparse 3D convolution network to generate a 3D volume, such that features at any sampled location (e.g., $\textbf{x}_i$) can be obtained through tri-linear interpolation. Similar ideas were used by~\cite{Peng2021neuralbody,Kwon21nhp}. Different from the prior works where the sparse convolution is implemented in the observation space, the SMPL feature diffusion process is implemented in the canonical space and serves as a category-level prior. 

\textbf{Volume Rendering.}  
As in NeRF~\cite{mildenhall2020nerf}, the expected rendering color $\mathbf{C(r)}$ along camera ray $\mathbf{r}$ is obtained by aggregating the predicted color $\mathbf{c}$ and density $\sigma$ generated by (\ref{eq: Q mapping}) using standard volume rendering. 

\subsection{Category-level Training}\label{method: training}
During training, given $N$ monocular videos sequences, in each iteration, ActorsNeRF randomly samples $K$ key frames from a monocular video and renders another sampled target image $I^t$ in the same video sequence. The image encoder $\mathcal{E}$ is used to extract corresponding features $\mathbf{U} = \{\mathbf{U}^1, \mathbf{U}^2, \cdots, \mathbf{U}^K \}$ for the $K$ key images. With these extracted image features, shown in Figure~\ref{fig: overview}, a ray of sampled points for a target rendering viewpoint $\mathbf{v}^t$ is aggregated to producing the corresponding color $\mathbf{C(r)}$. To ensure high-quality rendering, both the mean square error $\mathcal{L}_{mse}$ and the perceptual loss~\cite{zhang18lpips} $\mathcal{L}_{_{LPIPS}}$ are used as objective functions. Additionally, another skinning weight regularizer $ \mathcal{L}_{_{W}}$, which is an $\mathcal{L}_{1}$ loss, is employed to encourage the output skinning weights from the skinning weight network to be close to the prior obtained from the skeleton. All three objective functions are jointly optimized to update all network parameters,
\begin{equation}
    \mathcal{L} = \lambda_{mse} \mathcal{L}_{mse} + \lambda_{_{LPIPS}} \mathcal{L}_{_{LPIPS}} + \lambda_{_{W}} \mathcal{L}_{_{W}}
    \label{eq: loss}
\end{equation}
where $\lambda_{mse}$, $\lambda_{_{LPIPS}}$, and $\lambda_{_{W}}$ are corresponding coefficients to balance different loss functions.

\subsection{Few-shot Optimization}\label{method: few-shot optimization}
To transfer the knowledge for ActorsNerf learned at a category level to a novel human actor with $M$ frames provided, we propose to fine-tune the model to match the observations. During fine-tuning, we select $K$ out of $M$ frames as support frames for the encoder to extract features, such that all $M$ images are synthesized with these $K$ image features. Note that $M$ here can be much smaller compared to the category-level pretraining stage. Different from the training stage, where different $K$ frames are used for each iteration, the $K$ frames are fixed in the few-shot optimization stage for a stable performance in the few-shot setting. As the combination of the feature encoder and skinning weight network forms a category-level shape and appearance prior, only the deformation network and rendering network are fine-tuned, and the encoder and skinning weight network are frozen. Additionaly, only the mean square error and perceptual loss are used in the few-shot optimization stage. After fine-tuning, ActorsNeRF is capable of rendering the novel actor with novel viewpoints and poses using the $K$ support frames.

    \begin{figure*}[t]
        \centering
        \includegraphics[width=\linewidth]{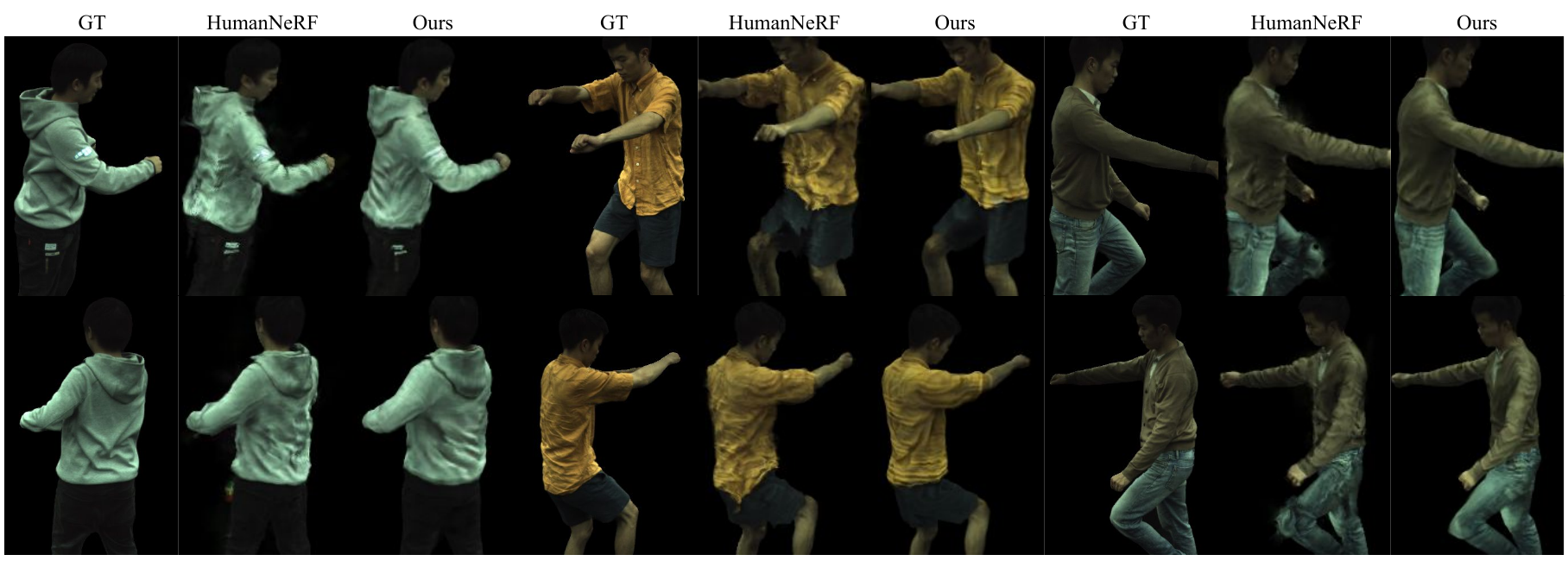}
        \vspace{-0.30in}
        \caption{Qualitative comparison for few-shot novel view synthesis of novel actors with unseen poses on the ZJU-MoCap dataset. Ground-truth (GT), HumanNeRF, and our results are shown from left to right. Top and bottom rows are rendered from different viewpoints.  Our method renders high-quality images with sharp boundaries and details.}
        \label{fig: 300frame-zju}
    \end{figure*}
    
    \begin{table*}[h]
    \fontsize{8}{10pt}\selectfont
    \setlength{\tabcolsep}{8pt}
    \centering
    \begin{tabular}{cc|ccc|ccc|ccc} 
    \cline{1-11}
    \multirow{2}{*}{} & & \multicolumn{3}{c|}{Person 387} & \multicolumn{3}{c|}{Person 393} & \multicolumn{3}{c}{Person 394} \\
    \cline{3-11}
    & & PSNR $\uparrow$ & SSIM $\uparrow$ & LPIPS $\downarrow$ & PSNR $\uparrow$ & SSIM $\uparrow$ & LPIPS $\downarrow$ & PSNR $\uparrow$ & SSIM $\uparrow$ & LPIPS $\downarrow$\\
    \shline
    \multirow{3}{*}{5-shot} & NeuralBody & 26.19 & 0.9404 & 85.91 & 27.54 & 0.9476 & 84.15 & 27.52 & 0.9453 & 82.20\\
    & HumanNeRF & 26.28 & 0.9465 & 65.64 & 26.84 & 0.9466 & 63.53 & 27.53 & 0.9451 & 64.15\\
    & Ours & \bf 27.26 & \bf 0.9568 & \bf 46.06 & \bf 27.20 & \bf 0.9553 & \bf 46.29 & \bf 28.09 & \bf 0.9577 & \bf 42.48\\
    \cline{1-11}
    \multirow{3}{*}{10-shot} & NeuralBody & 27.18 & 0.9494 & 76.32 & 27.17 & 0.9469 & 78.73 & 28.14 & 0.9492 & 77.23\\
    & HumanNeRF  & 26.66 & 0.9501 & 59.38 & 26.96 & 0.9516 & 56.29 & 28.51 & 0.9515 & 52.62\\
    & Ours & \bf 27.15 & \bf 0.9592 & \bf 40.89 & \bf 27.26 & \bf 0.9565 & \bf 42.56 & \bf 28.71 & \bf 0.9613 & \bf 35.93\\
    \cline{1-11}
    \multirow{3}{*}{30-shot} & NeuralBody & 26.32 & 0.9457 & 65.43 & 27.08 & 0.9482 & 72.85 & 28.10 & 0.9530 & 69.55\\
    & HumanNeRF  & 27.25 & 0.9555 & 47.59 & 27.37 & 0.9558 & 46.04 & 28.38 & 0.9559 & 43.76\\
    & Ours & \bf 27.67 & \bf 0.9610 & \bf 36.76 & \bf 27.59 & \bf 0.9577 & \bf 39.51 & \bf 28.97 & \bf 0.9614 & \bf 34.29\\
    \cline{1-11}
    \multirow{3}{*}{100-shot} & NeuralBody & 26.91 & 0.9513 & 60.60 & 27.13 & 0.9515 & 67.11 & 27.85 & 0.9544 & 59.42\\
    & HumanNeRF  & 27.13 & 0.9589 & 40.00 & 27.20 & 0.9550 & 42.91 & 28.25 & 0.9568 & 40.45\\
    & Ours & \bf 27.66 & \bf 0.9614 & \bf 36.39 & \bf 27.57 & \bf 0.9580 & \bf 39.33 & \bf 29.07 & \bf 0.9612 & \bf 34.03\\
    \cline{1-11}
    \multirow{3}{*}{300-shot} & NeuralBody & 26.95 & 0.9518 & 59.89 & 27.22 & 0.9535 & 64.60 & 27.69 & 0.9557 & 56.17 \\
    & HumanNeRF  & 27.30 & 0.9584 & 40.72 & 27.24 & 0.9557 & 43.59 & 28.46 & 0.9578 & 39.44\\
    & Ours & \bf 27.61 & \bf 0.9612 & \bf 36.18 & \bf 27.59 & \bf 0.9574 & \bf 39.36 & \bf 28.98 & \bf 0.9611 & \bf 34.17\\
    \cline{1-11}
    \end{tabular}
        \vspace{-0.10in}
    \caption{Few-shot generalization comparison for novel view synthesis of novel actors with unseen poses on the ZJU-MoCap dataset. }
    \label{table: 300frame-zju}
    \end{table*}

\section{Experiments}
We test ActorsNeRF on multiple benchmark datasets, e.g., ZJU-MoCap dataset~\cite{Peng2021neuralbody} and AIST++ dataset~\cite{Li21AIST}, and ActorsNeRF significantly outperforms multiple representative state-of-the-art baselines.


    \begin{figure*}[t]
        \centering
        \includegraphics[width=\linewidth]{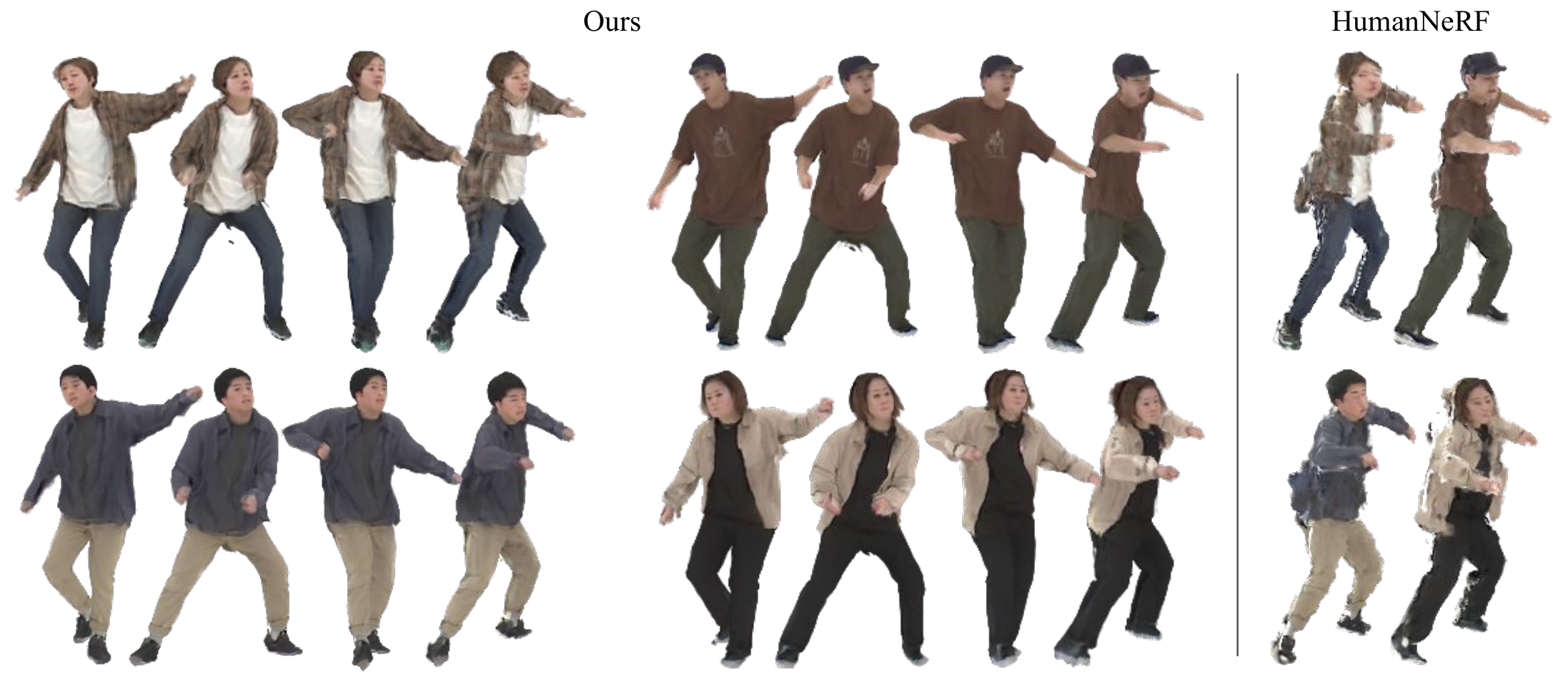}
        \vspace{-0.30in}
        \caption{Qualitative comparison for few-shot novel view synthesis of novel actors with unseen poses on the AIST++ dataset. Our method achieves high-quality animation with sharp boundary and details. In contrast, HumanNeRF outputs blurry images with broken body parts. }
        \label{fig: 300frame-AIST}
    \end{figure*}

    \begin{figure}[t]
        \centering
        \includegraphics[width=\linewidth]{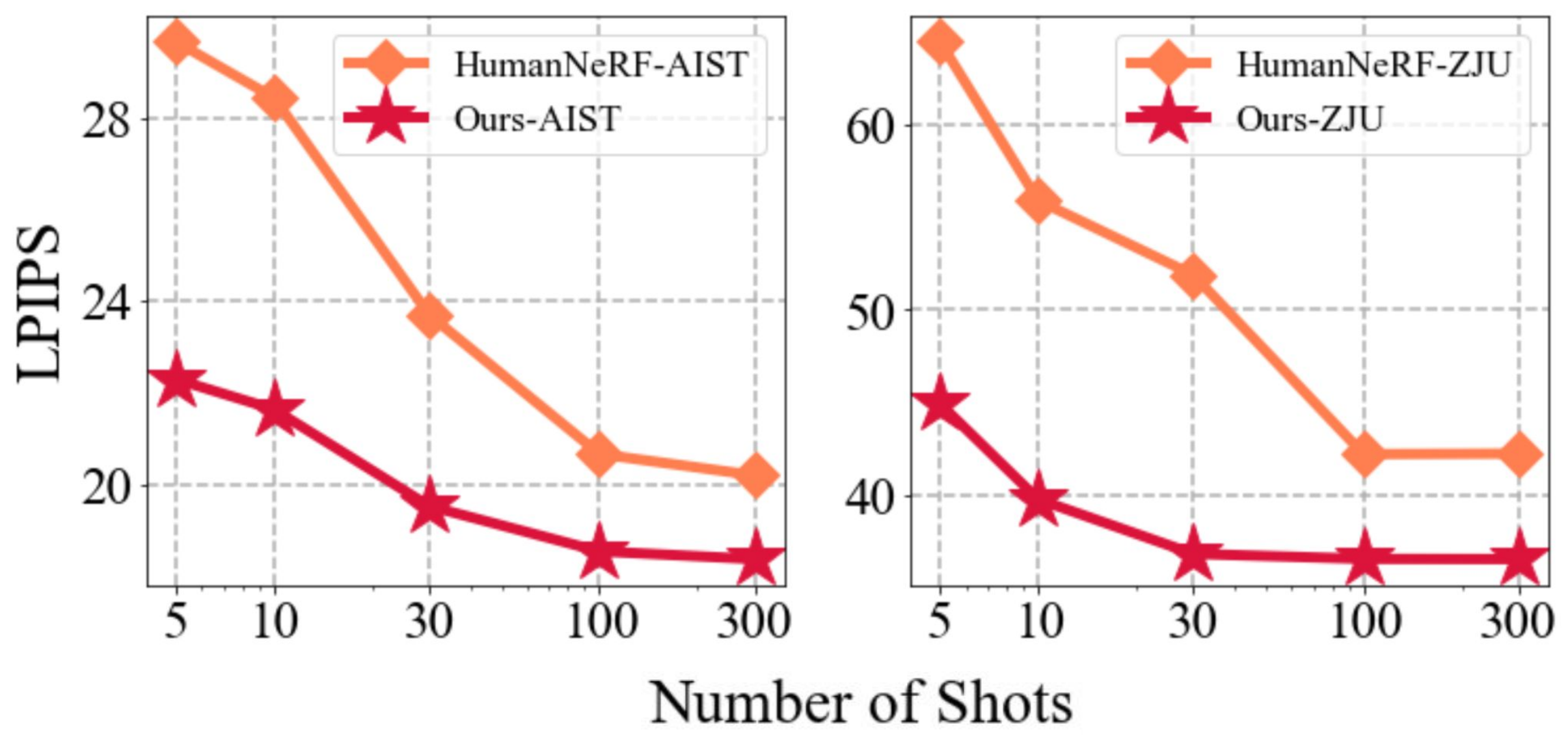}
        \vspace{-0.20in}
        \caption{Comparison for few-shot novel view synthesis of novel actors with unseen poses on ZJU-MoCap dataset (right) and AIST++ dataset (left). LPIPS scores are obtained by averaging across all test subjects in a dataset. }
        \label{fig: plot}
        \vspace{-0.25in}
    \end{figure}

    \begin{table*}[t]
    \fontsize{8}{11pt}\selectfont
    \setlength{\tabcolsep}{1.8pt}
    \centering
    \begin{tabular}{l c|ccc|ccc|ccc|ccc|ccc}
        \cline{1-17}
    \multirow{2}{*}{}& & \multicolumn{3}{c|}{Person 16} & \multicolumn{3}{c|}{Person 17} & \multicolumn{3}{c|}{Person 18} & \multicolumn{3}{c|}{Person 19} & \multicolumn{3}{c}{Person 20} \\
    \cline{3-17}
    & & PSNR $\uparrow$ & SSIM $\uparrow$ & LPIPS $\downarrow$ & PSNR $\uparrow$ & SSIM $\uparrow$ & LPIPS $\downarrow$ & PSNR $\uparrow$ & SSIM $\uparrow$ & LPIPS $\downarrow$ & PSNR $\uparrow$ & SSIM $\uparrow$ & LPIPS $\downarrow$ & PSNR $\uparrow$ & SSIM $\uparrow$ & LPIPS $\downarrow$ \\
    \shline
    \multirow{2}{*}{5-shot} & HN  & 24.37 & 0.9752 & 29.59 & 24.86 & 0.9762 & 29.39 & 22.77 & 0.9738 & 33.02 & 24.51 & 0.9759 & 28.68 & 24.55 & 0.9791 & 27.63\\
    & ours &\bf 25.22 & \bf 0.9796 & \bf 22.03 & \bf 25.88 & \bf 0.9808 & \bf 22.85 & \bf 24.50 & \bf 0.9811 & \bf 22.38 & \bf 25.24 & \bf 0.9801 & \bf 22.87 & \bf 25.30 & \bf 0.9827 & \bf 21.34\\
    \cline{1-17}
    \multirow{2}{*}{10-shot} & HN  & 24.22 & 0.9737 & 31.20 & 24.84 & 0.9758 & 28.81 & 23.96 & 0.9763 & 28.15 & 24.32 & 0.9758 & 28.86 & 24.94 & 0.9803 & 25.17\\
    & ours & \bf 25.25 & \bf 0.9794 & \bf 22.22 & \bf 25.87 & \bf 0.9812 & \bf 22.45 & \bf 24.71 & \bf 0.9822 & \bf 21.26 & \bf 25.33 & \bf 0.9809 & \bf 21.55 & \bf 25.53 & \bf 0.9833 & \bf 20.71\\
    \cline{1-17}
    \multirow{2}{*}{30-shot} & HN  & 25.08 & 0.9773 & 25.46 & 25.62 & 0.9793 & 24.06 & 24.53 & 0.9800 & 23.35 & 25.34 & 0.9793 & 23.77 & 25.41 & 0.9822 & 21.76\\
    & ours &\bf 25.67 & \bf 0.9806 & \bf 20.18 & \bf 26.06 & \bf 0.9826 & \bf 19.45 & \bf 24.82 & \bf 0.9826 & \bf 19.52 &\bf 25.53 & \bf 0.9818 & \bf 19.98 & \bf 25.58 & \bf 0.9840 & \bf 18.49\\
    \cline{1-17}
    \multirow{2}{*}{100-shot} & HN & 25.31 & 0.9783 & 21.87 & 25.81 & 0.9801 & 21.74 & 24.59 & 0.9811 & 20.84 & 25.69 & 0.9809 & 20.57 & 25.58 & 0.9837 & 18.29\\
    & ours & \bf 25.78 & \bf 0.9812 & \bf 19.05 & \bf 26.14 & \bf 0.9833 & \bf 18.47 & \bf 25.02 & \bf 0.9834 & \bf 18.64 & \bf 25.72 & \bf 0.9827 & \bf 18.84 & \bf 25.86 & \bf 0.9846 & \bf 17.69	\\
    \cline{1-17}
    \multirow{2}{*}{300-shot} & HN & 25.65 & 0.9795 & 21.54 & 26.12 & 0.9817 & 21.00 & 24.96 & 0.9824 & 19.98 & 25.79 & 0.9816 & 20.54 & \bf 26.00 & 0.9840 & 17.92\\
     & ours & \bf 25.73 & \bf 0.9812 & \bf 18.93	& \bf 26.14 & \bf 0.9834 & \bf 18.37 & \bf 25.03 & \bf 0.9833 & \bf 18.52 & \bf 25.88 & \bf 0.9827 & \bf 18.58 & 25.78 & \bf 0.9845 & \bf 17.44\\
    \cline{1-17}
    \end{tabular}
        \vspace{-0.10in}
    \caption{Few-shot generalization comparison for novel view synthesis of novel actors with unseen poses on the AIST++ dataset.}
        \vspace{-0.20in}
    \label{table: 300frame-AIST}
    \end{table*}

\subsection{Datasets and Baselines}\label{exp:dataset}
\textbf{Dataset.} We test ActorsNeRF on two datasets: ZJU-MoCap~\cite{Peng2021neuralbody} dataset and AIST++ dataset~\cite{Li21AIST}. The ZJU-MoCap dataset dataset contains 10 human subjects recorded from 21 / 23 multi-view cameras. We use the camera projections, body poses, and segmentations provided by the dataset. Follow ~\cite{Kwon21nhp}, We leave 3 (387, 393, 394) subjects as held-out data and use the remaining 7 for training. The AIST++ dataset is a dancing motion dataset capturing 30 human subjects performing various dances from 9 multi-view cameras. We randomly select one action sequence for each subject and then split the dataset with 25 actors for training and the other 5 actors (16-20) for testing. For both datasets, `camera1` is used for training and other views are only used for evaluation. More details are discussed in the supplementary materials.

    \begin{figure*}[t]
        \centering
        \includegraphics[width=\linewidth]{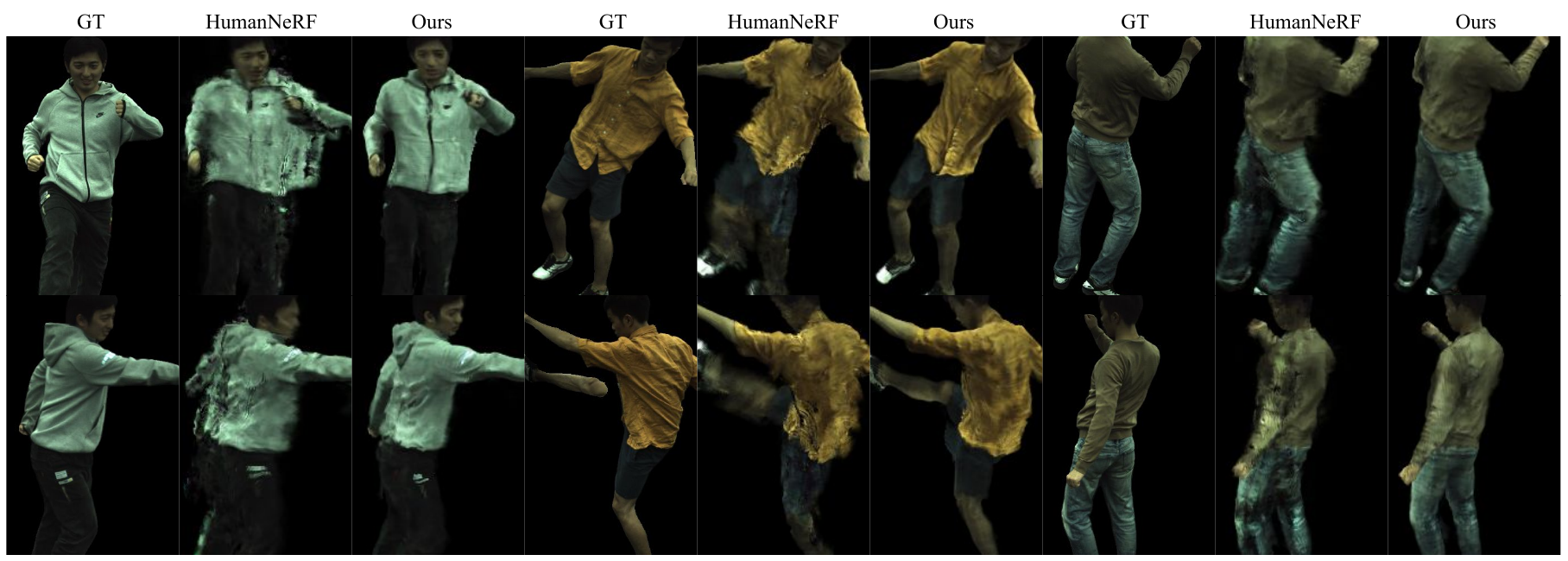}
        \vspace{-0.30in}
        \caption{Qualitative comparison for short-video novel view synthesis of novel actors with unseen poses on the ZJU-MoCap dataset. Each row presents a test actor with different poses. Ground-truth (GT), HumanNeRF, and our results are shown from left to right. ActorsNeRF generates sharp boundaries and maintains a better overall shape, indicating that the category level prior can be leveraged to smoothly synthesizing unobserved portions.}
        \label{fig: 100-shot}
    \end{figure*}

    \begin{table*}[t]
    \fontsize{8}{10pt}\selectfont
    \setlength{\tabcolsep}{8pt}
    \centering
    \begin{tabular}{l|ccc|ccc|ccc}
        \cline{1-10}
    \multirow{2}{*}{}& \multicolumn{3}{c|}{Person 387} & \multicolumn{3}{c|}{Person 393} & \multicolumn{3}{c}{Person 394} \\
    \cline{2-10}
    & PSNR $\uparrow$ & SSIM $\uparrow$ & LPIPS $\downarrow$ & PSNR $\uparrow$ & SSIM $\uparrow$ & LPIPS $\downarrow$ & PSNR $\uparrow$ & SSIM $\uparrow$ & LPIPS $\downarrow$\\
    \shline
    Humannerf  & 24.84 & 0.9352 & 81.82 & 25.76 & 0.9408 & 70.81 & 27.03 & 0.9460 & 65.82\\
    Ours & \bf 26.19 & \bf 0.9542 & \bf 50.88 & \bf 26.75 & \bf 0.9546 & \bf 47.02 & \bf 27.84 & \bf 0.9578 & \bf 44.77\\
        \cline{1-10}
    \end{tabular}
        \vspace{-0.10in}
    \caption{Short-video generalization comparison for novel view synthesis of novel actors with unseen poses on the ZJU-MoCap dataset.}
    \label{table: 100-shot}
    \end{table*}

    \begin{table*}[hbt!]
    \fontsize{8}{10pt}\selectfont
    \setlength{\tabcolsep}{6pt}
    \centering
    \begin{tabular}{l|ccc|ccc|ccc} 
        \cline{1-10}
    \multirow{2}{*}{}& \multicolumn{3}{c|}{Person 387} & \multicolumn{3}{c|}{Person 393} & \multicolumn{3}{c}{Person 394} \\
    \cline{2-10}
    & PSNR $\uparrow$ & SSIM $\uparrow$ & LPIPS $\downarrow$ & PSNR $\uparrow$ & SSIM $\uparrow$ & LPIPS $\downarrow$ & PSNR $\uparrow$ & SSIM $\uparrow$ & LPIPS $\downarrow$\\
    \shline
    Ours full model & 27.15 & 0.9592 & 40.89 & 27.26 & 0.9565 & 42.56 & 29.11 & 0.9613 & 35.93 \\
     w/out deformation network & 27.33 & 0.9587 & 41.12 & 27.32 & 0.9562 & 44.13 & 28.73 & 0.9591 & 36.61 \\
     w/out pixel-align local features & 27.06 & 0.9585 & 40.99 & 27.42 & 0.9569 & 42.77 & 28.96 & 0.9602 & 36.31 \\
     w/out SMPL local features & 27.32 & 0.9592 & 41.35 & 27.32 & 0.9567 & 43.34 & 28.89 & 0.9608 & 37.13\\
         \cline{1-10}
    \end{tabular}
        \vspace{-0.10in}
    \caption{Ablation on components of ActorsNeRF.}
        \vspace{-0.20in}
    \label{table: ablation}
    \end{table*}

\textbf{Baseline and Metric.} We compare our method with the most representative state-of-the-art view synthesis methods. HumanNeRF~\cite{Weng2022humannerf} (HN) aligns various poses in a canonical space and achieves state-of-the-art rendering performance from a monocular video. 
NeuralBody~\cite{Peng2021neuralbody} is a representative method for rendering from observation space. Neural Human Performer~\cite{Kwon21nhp} and MPS-NeRF~\cite{Gao22mpsnerf} require multi-view images for both training and inference so they are not directly comparable to our monocular setting. Following ~\cite{Weng2022humannerf}, we use three metrics for quantitative evaluation: peak signal-to-noise ratio (PSNR), structural similarity index (SSIM), and perceptual quality (LPIPS)~\cite{zhang18lpips} (reported by $\times 10^{3}$). More comparisons and training details are included in the supplementary materials.

\subsection{Generalization}\label{exp:generalization}
In this section, we analyze how ActorsNeRF generalizes in a few-shot setup. We demonstrate the generalization of the learned category-level prior in two settings: few-shot generalization in Section~\ref{exp:generalization-5shot} and short-video generalization in Section~\ref{exp:generalization-video}. The few-shot generalization setting samples images where the human actor is mostly observed (e.g., both front and back). In contrast, in the short-video generalization setting, the input frames are selected such that a subject is not densely covered. This experiment is designed in a way that the model needs to `imagine` the missing portions.

\subsubsection{Few-shot Generalization}\label{exp:generalization-5shot}

In this section, we vary the number of input images to test the novel view synthesis of novel actors with unseen poses. The support set consists of $m$ frames, where $m=\{ 5, 10, 30, 100, 300\}$, sampled uniformly from 300 consecutive frames of a monocular video of an unseen subject. The objective is to synthesize the actor from novel viewpoints with novel poses.

As shown in Table~\ref{table: 300frame-zju}, on the ZJU-MoCap dataset, ActorsNeRF outperforms all baseline methods by a large margin across all shots, especially on the LPIPS metric. Figure~\ref{fig: 300frame-zju} presents the 30-shot rendering with unseen poses of test actors in the ZJU-MoCap dataset from 2 different viewpoints. 
The rendering produced by ActorsNeRF is smoother and preserves high-quality details, while maintaining a valid shape with less body distortion. In comparison, HumanNeRF produces unsmooth, blurry textures with distorted boundaries in the rendered images.

To further demonstrate the effectiveness of the proposed method, we test our algorithm on the more challenging AIST++ dataset~\cite{Li21AIST}, which contains diverse subjects with complex dancing poses and loose clothes. We employ PointRend~\cite{Kirillov20pointrend} to obtain foreground masks automatically. This introduces additional challenges as the mask boundaries tend to be noisy. The quantitative results are shown in Table~\ref{table: 300frame-AIST}, where our method achieves a much better performance. Figure~\ref{fig: 300frame-AIST} visualizes how, on this dataset, HumanNeRF frequently fails to produce a valid shape (distorted bodies and broken body parts), whereas ActorsNeRF achieves a much better synthesis of the overall shapes and textures. This again demonstrates the effectiveness of the learned category-level prior.

We additionally plot the LPIPS metric, averaged on all test subjects, for different shots in Figure~\ref{fig: plot}. Our insight is that ActorsNeRF gains more from the learned category-level prior when inputting fewer images. However, noting that even when 300 frames are provided, ActorsNeRF still yields significant margin over the baseline. The results suggests that the category-level prior of ActorsNeRF improves the rendering quality over a large few-shot spectrum. More results and visualizations for all different shots are shown in the supplementary materials.

\subsubsection{Short-video Generalization}\label{exp:generalization-video}

In this section, we study whether our algorithm is capable of imagining missing portions of the body given only a handful of images of a new person, with the goal of rendering novel poses and viewpoints. Specifically, we consider a scenario where only a portion of the person is observed in all frames, which is a natural task for humans who can imagine the missing parts of the body. We examine whether our algorithm can perform similarly under these conditions, by sampling $m=10$ frames from 100 consecutive frames.

Table~\ref{table: 100-shot} shows that, on the ZJU-MoCap dataset, ActorsNeRF again outperforms HumanNeRF (HN) approach significantly. Figure~\ref{fig: 100-shot}  provides a clear visual comparison between different methods. HumanNeRF produces some details for the observed body portions, but fails to maintain a valid shape for the less unobserved body parts. In comparison, ActorsNeRF still generates sharp boundaries and keeps the overall shape, suggesting that the ActorsNeRF is capable of leveraging the category level prior to smoothly synthesize unobserved portions of the body.

\subsection{Ablation}\label{exp:ablation}
In this section, we investigate the efficacy of the 2-level canonical space and local encoder features of ActorsNeRF in the context of the 10-shot generalization setting on the ZJU-MoCap dataset. Specifically, we perform ablation experiments by excluding a component in both category-level training and instance-level fine-tuning. Our quantitative results are presented in Table~\ref{table: ablation}, while the qualitative results are provided in the supplementary materials.

As discussed in Section~\ref{method: ActorsNerf}, the deformation network is responsible for refining category-level shape into a fine-grained instance-level shape. We ablate the deformation network by directly rendering from the category-level canonical space, i.e., the rendering network directly takes as input the point in the category-level canonical space as input. As observed in Table~\ref{table: ablation}, the results indicate that the proposed 2-level canonical design is critical to rendering photorealistic details. Furthermore, we observe that removing either pixel-aligned local features or smpl local features results in lower LPIPS scores, indicating that all components are essential for achieving the best overall performance.

\section{Conclusion}

We introduce ActorsNeRF, a generalizable NeRF-based human representation, that is trained from monocular videos and adapted to novel human subjects with a few monocular images. ActorsNeRF encodes the category-level prior through parameter sharing on multiple human subjects, and is implemented with a 2-level canonical space to capture the large human appearance, shape, and pose variations. ActorsNeRF is tested on multiple benchmark datasets, e.g., ZJU-MoCap dataset and AIST++ dataset. Compared to existing state-of-the-art methods, ActorsNeRF demonstrates superior novel view synthesis performance for novel human actors with unseen poses across multiple few-shot settings.

{\small
\bibliographystyle{ieee_fullname}
\bibliography{egbib}

\begin{thebibliography}{10}\itemsep=-1pt

\bibitem{Casas14video}
Dan Casas, Marco Volino, John~P. Collomosse, and Adrian Hilton.
\newblock 4d video textures for interactive character appearance.
\newblock {\em Comput. Graph. Forum}, 33(2):371--380, 2014.

\bibitem{Chen21mvsnerf}
Anpei Chen, Zexiang Xu, Fuqiang Zhao, Xiaoshuai Zhang, Fanbo Xiang, Jingyi Yu,
  and Hao Su.
\newblock Mvsnerf: Fast generalizable radiance field reconstruction from
  multi-view stereo.
\newblock In {\em ICCV}, pages 14104--14113.

\bibitem{Chen22geoaug}
Di Chen, Yu Liu, Lianghua Huang, Bin Wang, and Pan Pan.
\newblock Geoaug: Data augmentation for few-shot nerf with geometry
  constraints.
\newblock In {\em ECCV}, pages 322--337, 2022.

\bibitem{Collet15highquality}
Alvaro Collet, Ming Chuang, Pat Sweeney, Don Gillett, Dennis Evseev, David
  Calabrese, Hugues Hoppe, Adam~G. Kirk, and Steve Sullivan.
\newblock High-quality streamable free-viewpoint video.
\newblock {\em {ACM} Trans. Graph.}, 34(4):69:1--69:13, 2015.

\bibitem{Corona22lisa}
Enric Corona, Tomas Hodan, Minh Vo, Francesc Moreno{-}Noguer, Chris Sweeney,
  Richard Newcombe, and Lingni Ma.
\newblock {LISA:} learning implicit shape and appearance of hands.
\newblock In {\em CVPR}, pages 20501--20511, 2022.

\bibitem{Deng22depthsupervised}
Kangle Deng, Andrew Liu, Jun{-}Yan Zhu, and Deva Ramanan.
\newblock Depth-supervised nerf: Fewer views and faster training for free.
\newblock In {\em CVPR}, pages 12872--12881, 2022.

\bibitem{Dou16fusion4d}
Mingsong Dou, Sameh Khamis, Yury Degtyarev, Philip~L. Davidson, Sean~Ryan
  Fanello, Adarsh Kowdle, Sergio Orts{-}Escolano, Christoph Rhemann, David Kim,
  Jonathan Taylor, Pushmeet Kohli, Vladimir Tankovich, and Shahram Izadi.
\newblock Fusion4d: real-time performance capture of challenging scenes.
\newblock {\em {ACM} Trans. Graph.}, 35(4):114:1--114:13, 2016.

\bibitem{Fang22TiNeuVox}
Jiemin Fang, Taoran Yi, Xinggang Wang, Lingxi Xie, Xiaopeng Zhang, Wenyu Liu,
  Matthias Nie{\ss}ner, and Qi Tian.
\newblock Fast dynamic radiance fields with time-aware neural voxels.
\newblock {\em arXiv preprint arXiv:2205.15285}, 2022.

\bibitem{Gao21dynamicviewsynthesis}
Chen Gao, Ayush Saraf, Johannes Kopf, and Jia{-}Bin Huang.
\newblock Dynamic view synthesis from dynamic monocular video.
\newblock In {\em ICCV}, pages 5692--5701, 2021.

\bibitem{Gao22mpsnerf}
Xiangjun Gao, Jiaolong Yang, Jongyoo Kim, Sida Peng, Zicheng Liu, and Xin Tong.
\newblock Mps-nerf: Generalizable 3d human rendering from multiview images.
\newblock {\em arXiv preprint arXiv:2203.16875}, 2022.

\bibitem{Grigorev19coordinate}
Artur Grigorev, Artem Sevastopolsky, Alexander Vakhitov, and Victor~S.
  Lempitsky.
\newblock Coordinate-based texture inpainting for pose-guided human image
  generation.
\newblock In {\em CVPR}, pages 12135--12144, 2019.

\bibitem{guo22neuraldeformablevoxel}
Xiang Guo, Guanying Chen, Yuchao Dai, Xiaoqing Ye, Jiadai Sun, Xiao Tan, and
  Errui Ding.
\newblock Neural deformable voxel grid for fast optimization of dynamic view
  synthesis.
\newblock {\em arXiv preprint arXiv:2206.07698}, 2022.

\bibitem{He16resnet}
Kaiming He, Xiangyu Zhang, Shaoqing Ren, and Jian Sun.
\newblock Deep residual learning for image recognition.
\newblock In {\em CVPR}, pages 770--778, 2016.

\bibitem{Jain21dietnerf}
Ajay Jain, Matthew Tancik, and Pieter Abbeel.
\newblock Putting nerf on a diet: Semantically consistent few-shot view
  synthesis.
\newblock In {\em ICCV}, pages 5865--5874, 2021.

\bibitem{Jiang22neuman}
Wei Jiang, Kwang~Moo Yi, Golnoosh Samei, Oncel Tuzel, and Anurag Ranjan.
\newblock Neuman: Neural human radiance field from a single video.
\newblock In {\em Proceedings of the European conference on computer vision
  (ECCV)}, 2022.

\bibitem{Kanade97virtualizedreality}
Takeo Kanade, Peter Rander, and P.~J. Narayanan.
\newblock Virtualized reality: Constructing virtual worlds from real scenes.
\newblock {\em {IEEE} Multim.}, 4(1):34--47, 1997.

\bibitem{Kania22conerf}
Kacper Kania, Kwang~Moo Yi, Marek Kowalski, Tomasz Trzcinski, and Andrea
  Tagliasacchi.
\newblock Conerf: Controllable neural radiance fields.
\newblock In {\em CVPR}, pages 18602--18611, 2022.

\bibitem{Kirillov20pointrend}
Alexander Kirillov, Yuxin Wu, Kaiming He, and Ross~B. Girshick.
\newblock Pointrend: Image segmentation as rendering.
\newblock In {\em CVPR}, pages 9796--9805, 2020.

\bibitem{Kwon21nhp}
Youngjoong Kwon, Dahun Kim, Duygu Ceylan, and Henry Fuchs.
\newblock Neural human performer: Learning generalizable radiance fields for
  human performance rendering.
\newblock In {\em NeurIPS}, pages 24741--24752, 2021.

\bibitem{Li22tava}
Ruilong Li, Julian Tanke, Minh Vo, Michael Zollh{\"{o}}fer, J{\"{u}}rgen Gall,
  Angjoo Kanazawa, and Christoph Lassner.
\newblock {TAVA:} template-free animatable volumetric actors.
\newblock {\em arXiv preprint arXiv:2206.08929}, 2022.

\bibitem{Li21AIST}
Ruilong Li, Shan Yang, David~A. Ross, and Angjoo Kanazawa.
\newblock {AI} choreographer: Music conditioned 3d dance generation with
  {AIST++}.
\newblock In {\em ICCV}, pages 13381--13392, 2021.

\bibitem{Li21neuralsceneflow}
Zhengqi Li, Simon Niklaus, Noah Snavely, and Oliver Wang.
\newblock Neural scene flow fields for space-time view synthesis of dynamic
  scenes.
\newblock In {\em CVPR}, pages 6498--6508, 2021.

\bibitem{Liu21neuralactor}
Lingjie Liu, Marc Habermann, Viktor Rudnev, Kripasindhu Sarkar, Jiatao Gu, and
  Christian Theobalt.
\newblock Neural actor: neural free-view synthesis of human actors with pose
  control.
\newblock {\em {ACM} Trans. Graph.}, 40(6):219:1--219:16, 2021.

\bibitem{Loper2015SMPL}
Matthew Loper, Naureen Mahmood, Javier Romero, Gerard Pons{-}Moll, and
  Michael~J. Black.
\newblock {SMPL:} a skinned multi-person linear model.
\newblock {\em {ACM} Trans. Graph.}, 34(6):248:1--248:16, 2015.

\bibitem{BruallaP18lookinggood}
Ricardo Martin{-}Brualla, Rohit Pandey, Shuoran Yang, Pavel Pidlypenskyi,
  Jonathan Taylor, Julien P.~C. Valentin, Sameh Khamis, Philip~L. Davidson,
  Anastasia Tkach, Peter Lincoln, Adarsh Kowdle, Christoph Rhemann, Dan~B.
  Goldman, Cem Keskin, Steven~M. Seitz, Shahram Izadi, and Sean~Ryan Fanello.
\newblock \emph{LookinGood}: enhancing performance capture with real-time
  neural re-rendering.
\newblock {\em {ACM} Trans. Graph.}, 37(6):255, 2018.

\bibitem{Matusik00imagebased}
Wojciech Matusik, Chris Buehler, Ramesh Raskar, Steven~J. Gortler, and Leonard
  McMillan.
\newblock Image-based visual hulls.
\newblock In Judith~R. Brown and Kurt Akeley, editors, {\em SIGGRAPH}, pages
  369--374. {ACM}, 2000.

\bibitem{mildenhall2020nerf}
Ben Mildenhall, Pratul~P Srinivasan, Matthew Tancik, Jonathan~T Barron, Ravi
  Ramamoorthi, and Ren Ng.
\newblock Nerf: Representing scenes as neural radiance fields for view
  synthesis.
\newblock {\em Communications of the ACM}, 65(1):99--106, 2021.

\bibitem{Niemeyer22regnerf}
Michael Niemeyer, Jonathan~T. Barron, Ben Mildenhall, Mehdi S.~M. Sajjadi,
  Andreas Geiger, and Noha Radwan.
\newblock Regnerf: Regularizing neural radiance fields for view synthesis from
  sparse inputs.
\newblock In {\em CVPR}, pages 5470--5480, 2022.

\bibitem{Noguchi21narf}
Atsuhiro Noguchi, Xiao Sun, Stephen Lin, and Tatsuya Harada.
\newblock Neural articulated radiance field.
\newblock In {\em ICCV}, pages 5742--5752, 2021.

\bibitem{Park21nerfie}
Keunhong Park, Utkarsh Sinha, Jonathan~T. Barron, Sofien Bouaziz, Dan~B.
  Goldman, Steven~M. Seitz, and Ricardo Martin{-}Brualla.
\newblock Nerfies: Deformable neural radiance fields.
\newblock In {\em ICCV}, pages 5845--5854, 2021.

\bibitem{Park21hypernerf}
Keunhong Park, Utkarsh Sinha, Peter Hedman, Jonathan~T. Barron, Sofien Bouaziz,
  Dan~B. Goldman, Ricardo Martin{-}Brualla, and Steven~M. Seitz.
\newblock Hypernerf: a higher-dimensional representation for topologically
  varying neural radiance fields.
\newblock {\em {ACM} Trans. Graph.}, 40(6):238:1--238:12, 2021.

\bibitem{Peng21animatablenerf}
Sida Peng, Junting Dong, Qianqian Wang, Shangzhan Zhang, Qing Shuai, Xiaowei
  Zhou, and Hujun Bao.
\newblock Animatable neural radiance fields for modeling dynamic human bodies.
\newblock In {\em ICCV}, pages 14294--14303, 2021.

\bibitem{Peng2021neuralbody}
Sida Peng, Yuanqing Zhang, Yinghao Xu, Qianqian Wang, Qing Shuai, Hujun Bao,
  and Xiaowei Zhou.
\newblock Neural body: Implicit neural representations with structured latent
  codes for novel view synthesis of dynamic humans.
\newblock In {\em CVPR}, pages 9054--9063, 2021.

\bibitem{Pumarola21dnerf}
Albert Pumarola, Enric Corona, Gerard Pons{-}Moll, and Francesc
  Moreno{-}Noguer.
\newblock D-nerf: Neural radiance fields for dynamic scenes.
\newblock In {\em CVPR}, pages 10318--10327, 2021.

\bibitem{Radford21clip}
Alec Radford, Jong~Wook Kim, Chris Hallacy, Aditya Ramesh, Gabriel Goh,
  Sandhini Agarwal, Girish Sastry, Amanda Askell, Pamela Mishkin, Jack Clark,
  Gretchen Krueger, and Ilya Sutskever.
\newblock Learning transferable visual models from natural language
  supervision.
\newblock {\em arXiv preprint arXiv:2103.00020}, 2021.

\bibitem{Raj21pva}
Amit Raj, Michael Zollh{\"{o}}fer, Tomas Simon, Jason~M. Saragih, Shunsuke
  Saito, James Hays, and Stephen Lombardi.
\newblock {PVA:} pixel-aligned volumetric avatars.
\newblock pages 11733--11742, 2021.

\bibitem{Roessle22densedepthprior}
Barbara Roessle, Jonathan~T. Barron, Ben Mildenhall, Pratul~P. Srinivasan, and
  Matthias Nie{\ss}ner.
\newblock Dense depth priors for neural radiance fields from sparse input
  views.
\newblock In {\em CVPR}, pages 12882--12891, 2022.

\bibitem{Sarkar21style}
Kripasindhu Sarkar, Vladislav Golyanik, Lingjie Liu, and Christian Theobalt.
\newblock Style and pose control for image synthesis of humans from a single
  monocular view.
\newblock {\em arXiv preprint arXiv:2102.11263}, 2021.

\bibitem{Sarkar21neuralrerender}
Kripasindhu Sarkar, Dushyant Mehta, Weipeng Xu, Vladislav Golyanik, and
  Christian Theobalt.
\newblock Neural re-rendering of humans from a single image.
\newblock {\em arXiv preprint arXiv:2101.04104}, 2021.

\bibitem{Shao22doublefield}
Ruizhi Shao, Hongwen Zhang, He Zhang, Mingjia Chen, Yanpei Cao, Tao Yu, and
  Yebin Liu.
\newblock Doublefield: Bridging the neural surface and radiance fields for
  high-fidelity human reconstruction and rendering.
\newblock In {\em CVPR}, pages 15851--15861, 2022.

\bibitem{su22danbo}
Shih{-}Yang Su, Timur~M. Bagautdinov, and Helge Rhodin.
\newblock {DANBO:} disentangled articulated neural body representations via
  graph neural networks.
\newblock {\em arXiv preprint arXiv:2205.01666}, 2022.

\bibitem{Su21anerf}
Shih{-}Yang Su, Frank Yu, Michael Zollh{\"{o}}fer, and Helge Rhodin.
\newblock A-nerf: Articulated neural radiance fields for learning human shape,
  appearance, and pose.
\newblock In {\em NeurIPS}, pages 12278--12291, 2021.

\bibitem{Tancik21learninit}
Matthew Tancik, Ben Mildenhall, Terrance Wang, Divi Schmidt, Pratul~P.
  Srinivasan, Jonathan~T. Barron, and Ren Ng.
\newblock Learned initializations for optimizing coordinate-based neural
  representations.
\newblock In {\em CVPR}, pages 2846--2855, 2021.

\bibitem{Wang21ibrnet}
Qianqian Wang, Zhicheng Wang, Kyle Genova, Pratul~P. Srinivasan, Howard Zhou,
  Jonathan~T. Barron, Ricardo Martin{-}Brualla, Noah Snavely, and Thomas~A.
  Funkhouser.
\newblock Ibrnet: Learning multi-view image-based rendering.
\newblock In {\em CVPR}, pages 4690--4699, 2021.

\bibitem{Wang22ARAH}
Shaofei Wang, Katja Schwarz, Andreas Geiger, and Siyu Tang.
\newblock {ARAH:} animatable volume rendering of articulated human sdfs.
\newblock {\em arXiv preprint arXiv:2210.10036}, 2022.

\bibitem{Wei22selfsupervised}
Fangyin Wei, Rohan Chabra, Lingni Ma, Christoph Lassner, Michael
  Zollh{\"{o}}fer, Szymon Rusinkiewicz, Chris Sweeney, Richard~A. Newcombe, and
  Mira Slavcheva.
\newblock Self-supervised neural articulated shape and appearance models.
\newblock In {\em CVPR}, pages 15795--15805, 2022.

\bibitem{Weng20vid2actor}
Chung{-}Yi Weng, Brian Curless, and Ira Kemelmacher{-}Shlizerman.
\newblock Vid2actor: Free-viewpoint animatable person synthesis from video in
  the wild.
\newblock {\em arXiv preprint arXiv:2012.12884}, 2020.

\bibitem{Weng2022humannerf}
Chung{-}Yi Weng, Brian Curless, Pratul~P. Srinivasan, Jonathan~T. Barron, and
  Ira Kemelmacher{-}Shlizerman.
\newblock Humannerf: Free-viewpoint rendering of moving people from monocular
  video.
\newblock In {\em CVPR}, pages 16189--16199, 2022.

\bibitem{Xian21spacetime}
Wenqi Xian, Jia{-}Bin Huang, Johannes Kopf, and Changil Kim.
\newblock Space-time neural irradiance fields for free-viewpoint video.
\newblock In {\em CVPR}, pages 9421--9431, 2021.

\bibitem{Xu21HNeRF}
Hongyi Xu, Thiemo Alldieck, and Cristian Sminchisescu.
\newblock H-nerf: Neural radiance fields for rendering and temporal
  reconstruction of humans in motion.
\newblock In Marc'Aurelio Ranzato, Alina Beygelzimer, Yann~N. Dauphin, Percy
  Liang, and Jennifer~Wortman Vaughan, editors, {\em NeurIPS}, pages
  14955--14966, 2021.

\bibitem{Yang22banmo}
Gengshan Yang, Minh Vo, Natalia Neverova, Deva Ramanan, Andrea Vedaldi, and
  Hanbyul Joo.
\newblock Banmo: Building animatable 3d neural models from many casual videos.
\newblock In {\em CVPR}, pages 2853--2863, 2022.

\bibitem{Ye13freeviewpoint}
Genzhi Ye, Yebin Liu, Yue Deng, Nils Hasler, Xiangyang Ji, Qionghai Dai, and
  Christian Theobalt.
\newblock Free-viewpoint video of human actors using multiple handheld kinects.
\newblock {\em {IEEE} Trans. Cybern.}, 43(5):1370--1382, 2013.

\bibitem{Yu2021pixelnerf}
Alex Yu, Vickie Ye, Matthew Tancik, and Angjoo Kanazawa.
\newblock pixelnerf: Neural radiance fields from one or few images.
\newblock In {\em CVPR}, pages 4578--4587, 2021.

\bibitem{zhang18lpips}
Richard Zhang, Phillip Isola, Alexei~A. Efros, Eli Shechtman, and Oliver Wang.
\newblock The unreasonable effectiveness of deep features as a perceptual
  metric.
\newblock In {\em CVPR}, pages 586--595, 2018.

\bibitem{Zhao22humannerf}
Fuqiang Zhao, Wei Yang, Jiakai Zhang, Pei Lin, Yingliang Zhang, Jingyi Yu, and
  Lan Xu.
\newblock Humannerf: Efficiently generated human radiance field from sparse
  inputs.
\newblock In {\em CVPR}, pages 7743--7753, 2022.

\end{thebibliography}
}

\end{document}